%% file: main.tex
\documentclass[sigconf,review,anonymous=false,review=false,nonacm]{acmart}

\usepackage{booktabs}
\usepackage{enumitem}

\setcopyright{none}
\renewcommand\footnotetextcopyrightpermission[1]{}

\newcommand{\E}{\mathbb{E}}
\newcommand{\Var}{\operatorname{Var}}
\newcommand{\Deltahat}{\widehat{\Delta}}
\newcommand{\method}{\textsc{Model-CV}}
\newtheorem{assumption}{Assumption}[section]

\title{Variance-Optimal Counterfactual Policy Design for Sample-Efficient Online A/B Testing}
\renewcommand{\shorttitle}{Variance-Optimal Online A/B Testing}

\author{Shashank Gupta}
\affiliation{
  \institution{Microsoft, India}
  \city{}
  \country{}
} 
\authornote{Most of this work was completed prior to joining Microsoft.}
\email{27392shashankgupta@gmail.com}

\author{Subhadeep Maji}
\authornote{Most of this work was completed prior to joining DeepL.}
\affiliation{%
  \institution{DeepL, UK}
  \country{}}
\email{subhadeepmaji@gmail.com}

\begin{document}

\begin{abstract}
\input{sections/abstract}

\end{abstract}


\maketitle

\input{sections/introduction}
\input{sections/related_work}

\input{sections/main_contribution}
\input{sections/experiments}
\input{sections/discussion}

\input{sections/conclusion}

\bibliographystyle{ACM-Reference-Format}
\bibliography{references}

\appendix
\input{sections/appendix}

\end{document}

%% file: sections/abstract.tex
Online A/B testing is the standard approach for evaluating changes to
user-facing systems, such as a new recommendation or ranking policy, before
deployment. While the difference-in-means estimator of a standard A/B test is
unbiased, it often suffers from high variance. When the treatment and production
policies choose the same actions with similar probabilities, the resulting
observations can add noise while providing little information about the
difference between the policies.
Consequently, A/B tests can require large amounts of traffic and long durations
to detect small treatment effects. Existing variance-reduction methods mainly improve
the estimator after the data-collection scheme has been fixed, and recent
overlap-aware designs only optimize a single traffic split between the two
policies; thereby, they cannot direct traffic to the actions on which the
policies disagree. 

In this paper, we jointly optimize how actions are sampled and how the resulting
observations are used to estimate the treatment effect. We make two technical contributions. First, building on
pairwise off-policy evaluation, we derive the variance-optimal action-level
logging policy for estimating the average treatment effect between two
policies, which has a closed-form solution for any fixed control variate.
Second, we use historical logs to jointly optimize the logging policy and a
control variate to further reduce variance. Our
experimental results in simulated policy A/B tests show that our designs
estimate the average treatment effect more accurately than existing variance reduction methods on every
evaluated setting, while generally improving statistical power and reducing how
often the worse policy is selected. As a result, our designs enable practitioners to reach reliable
deployment decisions with considerably less online traffic.

%% file: sections/introduction.tex
\section{Introduction}
Online controlled experiments are the standard way to evaluate changes to
user-facing systems~\cite{kohavi2020trustworthy}. Before deploying a new search,
recommendation, or ranking system, it is typically compared with the existing
production system in an online A/B test. The test estimates how the change
affects a business metric, and thereby provides evidence for the decision to
deploy it.

Many modern A/B tests compare policies: given a context, such as a user and a
request, a policy defines a probability distribution over the available actions,
e.g., over the items that could be shown to the user. We call the observed
outcome, such as whether the user clicks on the item, the reward. A conventional A/B test
first assigns each unit to the treatment or production policy, then samples an
action from the assigned policy. However, two policies may recommend many of the
same items, sometimes with the same probabilities, and an item that receives the
same probability under both policies contributes nothing to the difference
between their expected rewards~\cite{jeunen2026accelerating}. Nevertheless, both
experimental groups may show this item frequently, for instance, when the item is
popular, and the difference-in-means estimator includes the resulting reward
noise in both group averages. Pairwise off-policy estimation can use the known
action probabilities to account for this
overlap~\cite{jeunen2026accelerating,sakhi2026exploiting}.

\textbf{High variance and sample inefficiency.}
The difference-in-means estimator of a standard A/B test is unbiased, but its
variance determines how much traffic is needed to detect an effect~\cite{kohavi2020trustworthy}. This problem
is especially important for incremental policy updates: the action space can
contain tens or hundreds of actions, the two policies can overlap substantially,
and the true effect is often small. A high-variance test requires more
observations to attain a given power for the same effect size, which can delay
the deployment decision~\cite{kohavi2020trustworthy}.

\textbf{Limitations of existing variance reduction.}
Existing variance-reduction methods for online experiments mainly improve the
estimator after the experimental design has been fixed. Regression adjustment and
CUPED remove predictable variation from the outcome using pre-treatment
covariates~\cite{deng2013cuped,guo2021mlrate,poyarkov2016boosted}. Separately,
off-policy evaluation uses logged data to evaluate a
target policy from data collected by a different policy, known as the logging
policy~\cite{dudik2011doubly,horvitz1952generalization}. More recently,
researchers have used policy overlap to optimize traffic allocation between the
treatment and production policies~\cite{jeunen2026accelerating}. However,
adjusting this overall traffic split does not allow them to independently choose
how often each action is sampled in a given context. This restriction limits the
potential gains in sample efficiency. To the best of our knowledge, no
existing work has designed the action-level data collection of a policy A/B test
jointly with its estimator.

\textbf{Combining off-policy evaluation with A/B testing.}
We aim to estimate the difference between the expected rewards of a fixed
treatment policy and a fixed production policy---the average treatment effect
(ATE). Instead of first assigning each interaction to one of these policies, we
sample its action using a separate logging policy, which specifies how often
each action is sampled in a given context. We weight each observed reward by
the difference between the two policies' probabilities of choosing its action.
These weights also correct for how often the logging policy samples that action,
giving an estimate of the ATE~\cite{jeunen2024delta}. This approach
builds on pairwise off-policy evaluation (OPE), specifically
$\Delta$-IPS~\cite{jeunen2024delta}.
To reduce variance further, we use a control
variate~\cite{dudik2011doubly,gupta2024optimal} as a baseline reward prediction.
The estimator corrects this prediction using observed outcomes, so that
prediction errors do not bias the estimated effect. We jointly optimize the
logging policy and the control variate to minimize the variance of the ATE
estimator. We fit both on historical logs and fix both before collecting any new
data.

Our key result is a closed-form expression for the variance-optimal logging
policy under any fixed control variate. This logging policy samples every action
in proportion to the product of (i) the absolute difference between the two
policies' probabilities of that action and (ii) the reward uncertainty that
remains after applying the control variate. A balanced A/B test allocates equal
traffic to the two policies and samples actions according to the assigned policy.
Our optimal logging policy instead favors actions with larger probability
differences, especially when their remaining reward uncertainty is high.
For each candidate control variate, our closed-form result gives its
corresponding optimal sampling policy. We therefore optimize only the control
variate, rather than searching separately over both components; this is known
as partial minimization. We use either a scalar control variate or a model-based control variate whose reward prediction varies with
the context and action~\cite{dudik2011doubly}.

We evaluate our designs in simulated online A/B tests that capture the main
difficulties of incremental policy updates: stochastic policies over actions that
overlap broadly, reward noise that depends on the context and the action, and a
small treatment effect. We compare them with balanced A/B testing and methods
that optimize the traffic split between policies. We measure estimation accuracy,
statistical power, and how often the worse policy is selected. Across different
numbers of actions and amounts of data, our designs estimate the effect more
accurately than all baselines and, in most cases, detect it more often and select
the worse policy less often. In particular, our model-based control-variate
design reaches the mean squared error (MSE) of balanced A/B testing with 39\% to 51\% of the
evaluation traffic, reduces MSE relative to the stronger traffic-split baseline
by 26\% to 41\%, and, with 10,000 evaluation interactions, increases the
statistical power of balanced A/B testing by a factor of 1.5 to 2.1.

Our main contributions are:
\begin{itemize}
  \item A formulation of policy A/B testing as the joint design of an
        action-level logging policy and a control variate for estimating the
        ATE using a counterfactual estimator;
  \item A variance-optimal contextual logging policy for any fixed control
        variate in closed form, optimized via partial minimization, which allows
        us to eliminate the logging policy from the joint design problem;
  \item Scalar and model-based control-variate designs, together with a uniform
        safety backoff that retains a small probability of sampling every action,
        and a two-stage procedure that keeps the estimator
        unbiased when the historical reward model is misspecified; and
  \item Empirical evidence from simulated policy A/B tests that our designs
        improve estimation accuracy in all evaluated settings, and statistical
        power and policy selection in most settings, over balanced A/B testing
        and baselines that optimize the overall traffic split between policies.
\end{itemize}

%% file: sections/related_work.tex
\section{Related Work}

\textbf{Variance reduction in online experiments.}
Variance reduction has a long history in online experimentation.
CUPED uses pre-experiment covariates to remove predictable variation from the metric~\cite{deng2013cuped}, and subsequent work uses boosted trees and cross-fitted machine-learning models for the same purpose~\cite{poyarkov2016boosted,guo2021mlrate}.
These approaches improve the estimator, but keep the randomization scheme fixed.
We also use a historical reward model, but we treat the sampling policy as a design choice: the reward predictions help determine which actions to sample in the new experiment.

\textbf{Off-policy evaluation and control variates.}
Inverse propensity scoring (IPS) reweights observed rewards to evaluate a policy
from logs collected by another policy. It provides unbiased estimates under the
full support condition: the logger must have a positive probability of sampling
every action needed for evaluation~\cite{horvitz1952generalization,gupta2023safe,gupta2023recent,gupta2024unbiased,gupta2025safe,gupta2025towards}.
The doubly robust (DR) estimator combines reward predictions with an IPS
correction based on observed outcomes~\cite{dudik2011doubly}. Optimal OPE theory
studies the accuracy of estimators based on reward predictions, importance
weighting, or both~\cite{wang2017optimal}.
More robust doubly robust (MRDR) estimation fits the reward model to minimize the DR estimator's variance, rather than for outcome prediction alone~\cite{farajtabar2018more}.
Optimal baseline corrections give a closed-form constant baseline that minimizes
variance for a fixed logging policy~\cite{gupta2024optimal}.
$\Delta$-OPE estimates the treatment effect between two policies from data
collected by a single logging policy~\cite{jeunen2024delta}.
Our estimator builds on this two-policy comparison, while MRDR and optimal
baseline corrections motivate fitting the control variate to reduce variance.
We additionally optimize the logging policy jointly with the control variate.

\textbf{Data collection for policy evaluation.}
Several reinforcement-learning methods design data collection to estimate the
value of a single policy; the logger is called a behavior policy in this setting.
\citet{hanna2017data} derive and search for a behavior policy that minimizes the
mean squared error of importance-sampling evaluation. \citet{liu2023efficient}
derive a closed-form variance-optimal behavior policy and approximate it using offline data, while
\citet{liu2024doubly} jointly design the behavior policy and a baseline.
In contextual bandits, \citet{wan2022safe} study efficient, safe data collection to
evaluate and compare policies. \citet{tucker2023variance} design additional logging
to augment existing data for sample-efficient evaluation of a contextual-bandit policy.
We build on this work to estimate the treatment effect between two stochastic
policies, jointly optimizing the logger and a learned control variate and
evaluating power and policy selection.

\textbf{Adaptive experimental design.}
Neyman allocation assigns more traffic to treatments with more variable outcomes,
with probabilities proportional to their outcome standard deviations. Modern
adaptive designs extend this idea to treatment effects that vary across
contexts~\cite{kato2020efficient}.
Our result recovers the contextual version of this allocation when the two
policies always choose different actions and the control variate equals the true
mean reward. For stochastic policies, the allocation must also account for how
much their action probabilities differ.

\textbf{Policy overlap in A/B testing.}
\citet{jeunen2026accelerating} shows that overlap between the treatment and production policies can reduce the variance of policy A/B tests.
The method optimizes one overall traffic split between the two policies, rather
than choosing each action's sampling probability independently. This is a
policy-mixture design, which we use as our main overlap-aware baseline.
\citet{jeunen2026accelerating} also considers model-based control variates through
the pairwise MRDR estimator, $\Delta$-MRDR.
Other work improves estimation within an A/B test. \citet{sakhi2026exploiting} use
the policies' action probabilities, changing only the estimator, not data
collection. \citet{konishi2026more} reduce algorithm-selection errors by using a
hypothetical intermediate algorithm to break the original comparison into
separate steps.
In contrast, we also change data collection: for each context, our logger sets
individual action probabilities using the differences between the policies and
the reward uncertainty remaining after control-variate correction, rather than
only an overall traffic split.

%% file: sections/main_contribution.tex
\section{Variance-Optimal Policy A/B Testing}
\label{sec:method}

We formulate policy A/B testing as pairwise off-policy evaluation, introduce a control-variate estimator, and derive its variance-optimal logger.
Eliminating the logger by partial minimization yields our scalar and model-based designs.

\subsection{Problem setup}

We consider a contextual decision problem with contexts $X\in\mathcal X$, a finite action set $\mathcal A=\{1,\ldots,K\}$, and rewards $Y\in[0,1]$. Our theory holds for any bounded reward; for simplicity, our estimation procedure and experiments use binary rewards.
A stochastic policy $\pi$ chooses action $a$ in context $x$ with probability $\pi(a\mid x)$, and its value is its expected reward:
\begin{align}
 V(\pi)&=\E_X\left[\sum_{a\in\mathcal A}
 \pi(a\mid X)\mu(X,a)\right], \nonumber \\
 \mu(x,a)&=\E[Y\mid X=x,A=a].
\end{align}
We aim to compare a treatment policy $\pi_t$ with a production policy $\pi_p$.
The average treatment effect (ATE) of deploying $\pi_t$ instead of $\pi_p$ is
\begin{align}
 \Delta&=V(\pi_t)-V(\pi_p)
 =\E_X\left[\sum_a d(X,a)\mu(X,a)\right], \nonumber\\
 d(x,a)&=\pi_t(a\mid x)-\pi_p(a\mid x).
 \label{eq:pairwise-ate}
\end{align}
We refer to $d(x,a)$ as the \emph{policy contrast} of action $a$ in context $x$.
Both policies are available before the experiment, so the contrast is known and sums to zero: $\sum_a d(x,a)=0$ for every context.

During the experiment, actions are sampled from a contextual \emph{logging policy} $q(a\mid x)$, which we also call the \emph{logger}.
For $n$ independent units, we observe $\mathcal D=\{(X_i,A_i,Y_i)\}_{i=1}^n$, where $A_i\sim q(\cdot\mid X_i)$.
We fit the logger and control variate on historical logs, separate from the experiment's \emph{evaluation data}, under the following assumption.

\begin{assumption}[Identification and two-stage design]
Contexts are sampled independently from the target population; the conditional reward distribution is the same in the historical and evaluation stages; and the logger satisfies $q(a\mid x)>0$ whenever $d(x,a)\neq0$.
The logger and all control-variate parameters are fitted only on an independent historical dataset and frozen before observing any evaluation outcomes.
\label{ass:design}
\end{assumption}

The pairwise inverse propensity scoring (IPS) estimator of the ATE is:
\begin{equation}
 \widehat\Delta_{\mathrm{IPS}}
 =\frac{1}{n}\sum_{i=1}^n
 \frac{d(X_i,A_i)}{q(A_i\mid X_i)}Y_i.
 \label{eq:pairwise-ips}
\end{equation}
Introduced as $\Delta$-IPS by~\citet{jeunen2024delta}, it is unbiased under Assumption 3.1:
\begin{align}
 \E_{A,Y\mid X=x}\left[
 \frac{d(x,A)}{q(A\mid x)}Y\right]
 &=\sum_a q(a\mid x)\frac{d(x,a)}{q(a\mid x)}\mu(x,a)
 \nonumber\\
 &=\sum_a d(x,a)\mu(x,a).
 \label{eq:ips-unbiased}
\end{align}
Eq.~\eqref{eq:pairwise-ips} exploits policy overlap: actions with $d(x,a)=0$ receive zero weight.
Its variance can still be large when $q(a\mid x)$ is small, or rewards are noisy.
We address both sources by jointly choosing the control variate and logger.

\subsection{A pairwise control-variate estimator}
Let $m(x,a)$ be the control variate, either estimated from historical data or set to a constant.
We define the predicted policy effect $g_m(x)$ as:
\begin{equation}
 g_m(x)=\sum_a d(x,a)m(x,a),
 \label{eq:direct-term}
\end{equation}
and the per-observation score as:
\begin{equation}
 Z_m(x,a,y)=g_m(x)+
 \frac{d(x,a)}{q(a\mid x)}\{y-m(x,a)\}.
 \label{eq:score}
\end{equation}
The resulting estimator is $\widehat\Delta_m=n^{-1}\sum_{i=1}^n Z_m(X_i,A_i,Y_i)$.
The estimator combines the model's predicted policy effect with an importance-weighted reward correction. For $m \equiv 0$, this reduces to the IPS estimator in Eq.~(3); for a constant
control variate $m(x,a)=\beta$, it reduces to the $\Delta_\beta$-IPS
estimator~\cite{jeunen2024delta}; and for a reward model, it is a pairwise
doubly robust estimator~\cite{dudik2011doubly,jeunen2026accelerating}. In all
cases, the correction preserves unbiasedness even when $m$ differs from the true
reward mean.

\begin{proposition}[Unbiasedness]
Under Assumption~\ref{ass:design}, $\E[\widehat\Delta_m]=\Delta$ for every function $m$ that is fitted on historical data and every logging policy $q$ that satisfies the support condition.
\label{prop:unbiased}
\end{proposition}

\begin{proof}
Conditioning on the context and on the frozen historical data gives
\begin{align}
 \E[Z_m\mid X=x]
 &=g_m(x)+\sum_a d(x,a)\{\mu(x,a)-m(x,a)\}
 \nonumber\\
 &=\sum_a d(x,a)\mu(x,a).
 \label{eq:unbiased}
\end{align}
Taking the expectation over $X$ gives Eq.~\eqref{eq:pairwise-ate}.
\end{proof}

\subsection{Variance of the pairwise estimator}

Let $\tau(x)$ denote the conditional treatment effect and $s_m(x,a)$ the residual reward's second moment:
\begin{equation}
 \tau(x)=\sum_a d(x,a)\mu(x,a),
 \;\;
 s_m(x,a)=\E[(Y-m(x,a))^2\mid x,a].
 \label{eq:tau-sm}
\end{equation}
For Bernoulli rewards,
\begin{equation}
 s_m(x,a)=\mu(x,a)\{1-\mu(x,a)\}
 +\{\mu(x,a)-m(x,a)\}^2.
 \label{eq:bernoulli-sm}
\end{equation}
Expanding Eq.~\eqref{eq:score} gives
\begin{align}
 \E[Z_m^2\mid X=x]
 ={}&\sum_a\frac{d(x,a)^2s_m(x,a)}{q(a\mid x)}
 \nonumber\\
 &+2g_m(x)\tau(x)-g_m(x)^2.
 \label{eq:conditional-second-moment}
\end{align}
Therefore, the exact variance of a single score is
\begin{align}
 \mathcal V(m,q)=\E_X\bigg[&
 \sum_a\frac{d(X,a)^2s_m(X,a)}{q(a\mid X)}
 \nonumber\\
 &+2g_m(X)\tau(X)-g_m(X)^2\bigg]-\Delta^2,
 \label{eq:exactvariance}
\end{align}
and $\Var(\widehat\Delta_m)=\mathcal V(m,q)/n$. The logging policy affects only the first term. The control variate affects the residual rewards in this term and the correction terms that follow. We therefore fit the control variate by directly minimizing the estimator’s variance.

\subsection{Optimal logger for a fixed control variate}

For fixed $m$ and $x$, only the first term of Eq.~\eqref{eq:exactvariance} depends on $q$. The contextual design problem is therefore:
\begin{equation}
 \min_{q(\cdot\mid x)}
 \sum_a\frac{d(x,a)^2s_m(x,a)}{q(a\mid x)}
 \quad\text{s.t.}\quad
 \sum_a q(a\mid x)=1,\;q(a\mid x)\geq0.
 \label{eq:loggerobjective}
\end{equation}

\begin{theorem}[Variance-optimal contextual logger]
For a fixed control variate $m$ satisfying $s_m(x,a)>0$ whenever $d(x,a)\neq0$, the optimal logging policy at every context where the policies differ is:
\begin{equation}
 q_m^*(a\mid x)=
 \frac{|d(x,a)|\sqrt{s_m(x,a)}}
 {\sum_b|d(x,b)|\sqrt{s_m(x,b)}}.
 \label{eq:optimalq}
\end{equation}
If $d(x,a)=0$ for every action, the objective is zero for any logger; we define $q_m^*(a\mid x)=1/K$ in this case. Terms with $d(x,a)=0$ are understood to contribute zero. Under these conditions, $q_m^*$ minimizes the variance of $\widehat\Delta_m$ among all contextual logging policies that satisfy the support condition.
\label{thm:optimal-logger}
\end{theorem}

\begin{proof}
Consider a context where the policies differ and let $c_a=d(x,a)^2s_m(x,a)$.
The Lagrangian for Eq.~\eqref{eq:loggerobjective} is
\begin{equation}
 \mathcal L(q,\lambda)=\sum_a\frac{c_a}{q_a}
 +\lambda\left(\sum_a q_a-1\right).
\end{equation}
For every action with $c_a>0$, the first-order condition is $-c_a/q_a^2+\lambda=0$, and hence $q_a=\sqrt{c_a/\lambda}$; actions with $c_a=0$ do not contribute to the objective and receive zero probability.
Normalizing over the actions gives $q_a=\sqrt{c_a}/\sum_b\sqrt{c_b}$, which is exactly Eq.~\eqref{eq:optimalq}.
The objective is convex in $q$ on the supported simplex, so this stationary point is the global minimum.
The terms of Eq.~\eqref{eq:exactvariance} outside the summation do not depend on $q$.
\end{proof}

\begin{corollary}[Dominance over policy mixtures]
Under the conditions of Theorem~\ref{thm:optimal-logger}, for a fixed control variate $m$ and context $x$, let $C_m(q\mid x)$ denote the objective in Eq.~\eqref{eq:loggerobjective}.
Then, for every supported logger $q$,
\begin{align}
 C_m(q\mid x)&\geq
 \left\{\sum_a |d(x,a)|\sqrt{s_m(x,a)}\right\}^2 \nonumber\\
 &=C_m(q_m^*\mid x).
 \label{eq:logger-dominance}
\end{align}
In particular, the inequality holds for balanced A/B and for every supported logger in the policy-mixture family $q_\rho=\rho\pi_t+(1-\rho)\pi_p$~\cite{jeunen2026accelerating}.
\label{cor:dominance}
\end{corollary}

\begin{proof}
Applying the Cauchy--Schwarz inequality to the factors $\sqrt{q(a\mid x)}$ and $|d(x,a)|\sqrt{s_m(x,a)/q(a\mid x)}$ gives Eq.~\eqref{eq:logger-dominance}, with equality when $q=q_m^*$.
\end{proof}

Intuitively, the optimal logger favors actions on which the policies disagree, because only these actions carry information about the difference between the policies.
It also favors actions with greater residual reward uncertainty, which would otherwise contribute more variance.
A policy mixture generally cannot represent this allocation: changing the treatment share $\rho$ changes all action probabilities through the same mixing weight, so it cannot target individual actions.

Corollary~\ref{cor:dominance} shows that, for a fixed control variate, balanced A/B is generally suboptimal. Its logging policy,
$q_{\mathrm{AB}}(a\mid x)=\{\pi_t(a\mid x)+\pi_p(a\mid x)\}/2$,
can assign substantial traffic to actions that both policies
choose with nearly equal probabilities, yet these actions contribute little to the difference between the
policies. 

The choice of control variate determines how much reward
variation remains. 
For example, if $m=\mu$, then $s_m(x,a)=\mathrm{Var}(Y\mid x,a)$ and Eq.~(13)
allocates traffic in proportion to $|d(x,a)|\sqrt{\mathrm{Var}(Y\mid x,a)}$;
when the policies deterministically choose two different actions, this is
contextual Neyman allocation. We next describe how to choose the control variate
and the logging policy jointly.

\subsection{Joint optimization of logger and control variate}

Let $\mathcal M$ be the chosen class of control variates and $\mathcal Q$ the set of contextual loggers that satisfy the support condition.
We jointly choose $m$ and $q$ in two steps.
First, fix $m$ and choose its variance-minimizing logger:
\begin{equation}
 q_m^*=\arg\min_{q\in\mathcal Q}\mathcal V(m,q).
 \label{eq:joint-design}
\end{equation}
Theorem~\ref{thm:optimal-logger} gives this logger in closed form.
Next, minimize the resulting variance $\mathcal F(m)=\mathcal V(m,q_m^*)$ over the control-variate class:
\begin{equation}
 m^*=\arg\min_{m\in\mathcal M}\mathcal V(m,q_m^*)
     =\arg\min_{m\in\mathcal M}\mathcal F(m).
 \label{eq:partial-minimization}
\end{equation}
The final design is the pair $(m^*,q_{m^*}^*)$.
Eliminating the logger yields an exact population objective. 

Substituting the optimal logger from Eq.~\eqref{eq:optimalq}
into Eq.~\eqref{eq:exactvariance} gives an objective that
depends only on the control variate:
\begin{align}
 \mathcal F(m)=\E_X\bigg[&
 \left\{\sum_a |d(X,a)|\sqrt{s_m(X,a)}\right\}^2
 \nonumber\\
 &+2g_m(X)\tau(X)-g_m(X)^2\bigg]-\Delta^2.
 \label{eq:profile}
\end{align}
Thus, instead of optimizing one probability vector per context, we optimize $m$; its logger follows from Eq.~\eqref{eq:optimalq}. In practice, we numerically optimize an estimated, backoff-adjusted objective, without guaranteeing convergence to the global optimum. 

\paragraph{Estimating the objective from historical data.}
We cannot evaluate Eq.~\eqref{eq:profile} directly because
$s_m(x,a)$ and $\tau(x)$ depend on the unknown reward mean
$\mu(x,a)$. We therefore fit a reward model $\widehat\mu(x,a)$
on historical logs. For Bernoulli rewards, we estimate these
quantities as:
\begin{align}
 \widehat s_m(x,a)
 &=\widehat\mu(x,a)\{1-m(x,a)\}^2 \nonumber\\
 &\quad+\{1-\widehat\mu(x,a)\}m(x,a)^2, \nonumber\\
 \widehat\tau(x)
 &=\sum_a d(x,a)\widehat\mu(x,a).
 \label{eq:plugin-quantities}
\end{align}
We substitute these estimates into Eq.~\eqref{eq:profile}
and average over the historical contexts to obtain the
objective used to optimize $m$. The reward model
$\widehat\mu$ remains fixed during this optimization;
only the control variate $m$ is optimized.

\subsection{Control variate design}
We consider scalar and model-based control variates.

\textbf{Scalar control variate.}
We first consider a constant control variate, $m(x,a)=\beta$.
For a fixed logger, setting the derivative of Eq.~\eqref{eq:exactvariance} with respect to $\beta$ to zero gives the optimal baseline
\begin{equation}
 \beta^*(q)=
 \frac{\E_X\sum_a d(X,a)^2\mu(X,a)/q(a\mid X)}
 {\E_X\sum_a d(X,a)^2/q(a\mid X)}.
 \label{eq:beta}
\end{equation}
The Overlap-aware A/B baseline uses this fixed-logger optimum (Section~\ref{sec:baselines}).
In contrast, after eliminating the logger, we directly minimize the profiled objective over the scalar $\beta$.
Let $d_{ia}=d(x_i,a)$ and $\widehat\mu_{ia}=\widehat\mu(x_i,a)$ for the $i$-th of the $n_{\mathrm{hist}}$ historical contexts.
For Bernoulli rewards, the historical-log estimate of the profiled objective is
\begin{align}
 \widehat{\mathcal F}(\beta)=\frac{1}{n_{\mathrm{hist}}}\sum_{i=1}^{n_{\mathrm{hist}}}
 \bigg\{\sum_a |d_{ia}|\big[&\widehat\mu_{ia}(1-\beta)^2
 \nonumber\\
 &+(1-\widehat\mu_{ia})\beta^2\big]^{1/2}\bigg\}^2,
 \label{eq:scalarprofile}
\end{align}
up to terms that do not depend on $\beta$.
In practice, we minimize the backoff-aware analog of this objective, i.e.,
Eq.~(28) with $m(x,a)=\beta$, over the bounded interval $[0,1]$.

\textbf{Model-based control variate.}
Whereas a scalar gives every context and action the same correction, a model $m_\theta(x,a)$ captures systematic variation across contexts and actions.
In our experiments, we use action-wise logistic regression:
\begin{equation}
 m_\theta(x,a)=\sigma(w_a^\top x+b_a),
 \label{eq:sigmoid}
\end{equation}
where $\sigma(u)=1/(1+e^{-u})$, and every action $a$ has its own slope vector $w_a$ and intercept $b_a$.
The sigmoid keeps $m_\theta$ in $[0,1]$, so that the unweighted residual $Y-m_\theta(X,A)$ lies in $[-1,1]$ for Bernoulli rewards, although the importance weight can still amplify this residual.
We call the resulting method the \emph{model-based control variate}, or \method{}.
We use logistic regression in this paper; it is not a restriction of the general estimator.

To fit the logistic model by minimizing estimated variance, write
$m_{ia}(\theta)=m_\theta(x_i,a)$ and
$\widehat s_{ia}(\theta)=\widehat s_{m_\theta}(x_i,a)$, using the residual
second moment in Eq.~\eqref{eq:plugin-quantities}. For historical context $x_i$, define
\begin{align}
 R_i(\theta)&=\sum_a|d_{ia}|\sqrt{\widehat s_{ia}(\theta)}, \nonumber\\
 g_i(\theta)&=\sum_a d_{ia}m_{ia}(\theta).
 \label{eq:model-quantities}
\end{align}
Here, $\widehat\mu_{ia}$ is fixed after the historical reward model has been fitted, and $\widehat\tau_i=\sum_a d_{ia}\widehat\mu_{ia}$ is the estimated conditional treatment effect.
Substituting these quantities into the profiled objective in Eq.~\eqref{eq:profile} gives the empirical objective
\begin{equation}
 \widehat{\mathcal F}(\theta)=\frac{1}{n_{\mathrm{hist}}}
 \sum_{i=1}^{n_{\mathrm{hist}}}
 \left[R_i(\theta)^2+2g_i(\theta)\widehat\tau_i-g_i(\theta)^2\right],
 \label{eq:model-objective}
\end{equation}
where we omit the constant $-\Delta^2$, as it does not affect the minimizer.
For any value of $\theta$, we obtain the matching variance-optimal logger without a second optimization:
\begin{equation}
 q_\theta^*(a\mid x_i)=
 \frac{|d_{ia}|\sqrt{\widehat s_{ia}(\theta)}}{R_i(\theta)}.
 \label{eq:model-logger}
\end{equation}
Eq.~\eqref{eq:model-objective} is unconstrained in $\theta$ alone, avoiding alternation between model fitting and constrained logger optimization.

\textbf{How the model reduces variance.}
In Eq.~\eqref{eq:score}, we apply importance weighting to
the residual $Y-m_\theta(x,a)$ rather than to the reward
itself. The model-based term $g_{m_\theta}(x)$ accounts
for the part subtracted from the reward, preserving the
expected ATE. To understand the remaining variation,
we can write
\begin{align}
 s_{m_\theta}(x,a)
 &=\E[(Y-m_\theta(x,a))^2\mid x,a] \nonumber\\
 &=\Var(Y\mid x,a)
   +\{\mu(x,a)-m_\theta(x,a)\}^2.
 \label{eq:residual-decomposition}
\end{align}
The first term is irreducible reward variation; the second
is the control variate's squared error relative to the
expected reward, which vanishes when $m_\theta=\mu$.

If the class of control variates were unrestricted, the minimizer of Eq.~(24)
would be the reward model itself, $m=\hat{\mu}$: since
$\sum_a q(a\mid x)=1$, the Cauchy--Schwarz inequality gives
$\sum_a d(x,a)^2\{\hat{\mu}(x,a)-m(x,a)\}^2/q(a\mid x)
\geq \{\hat{\tau}(x)-g_m(x)\}^2$, so any deviation from $\hat{\mu}$ increases the
estimated variance. A bounded, low-capacity model such as Eq.~(22) cannot
represent $\hat{\mu}$ exactly. Minimizing Eq.~(24) then selects the approximation
most relevant to the variance of the ATE estimator: it weights errors by
$d(x,a)^2/q(a\mid x)$, so that errors on different actions do not contribute
equally, and it accounts for the induced logger and the model-based policy
difference $g_{m_\theta}(x)$. In contrast, a standard prediction loss weights all
contexts and actions equally. We therefore fit the model using the full
objective in Eq.~(24), rather than a standard prediction loss.

The logging policy is updated together with the model
using Eq.~\eqref{eq:model-logger}. It allocates traffic
according to both the policy contrast and the estimated
residual second moment. Thus, the model adjusts the
rewards used in the estimator, while the logger adjusts
how often each action is sampled.

\textbf{Safety backoff.}
To avoid assigning very small probabilities to actions whose residual uncertainty is underestimated,
we mix every learned contextual logger with uniform exploration:
\begin{equation}
 q_{\mathrm{safe}}(a\mid x)
 =(1-\varepsilon)q_m^*(a\mid x)+\varepsilon/K,
 \qquad \varepsilon=0.05.
 \label{eq:backoff}
\end{equation}
This guarantees that $q_{\mathrm{safe}}(a\mid x)\geq0.05/K$. 

\textbf{Practical optimization.}
We account for the backoff during fitting by minimizing the variance under the backoff-adjusted logger $q_{\mathrm{safe},\theta}$, obtained by applying  Eq.~\eqref{eq:backoff} to Eq.~\eqref{eq:model-logger}:
\begin{align}
 \widehat{\mathcal F}_{\mathrm{safe}}(\theta)
 =\frac{1}{n_{\mathrm{hist}}}\sum_i\bigg[&
 \sum_a\frac{d_{ia}^2\widehat s_{ia}(\theta)}
 {q_{\mathrm{safe},\theta}(a\mid x_i)}
 \nonumber\\
 &+2g_i(\theta)\widehat\tau_i-g_i(\theta)^2\bigg].
 \label{eq:safe-model-objective}
\end{align}
When $\varepsilon=0$, the summation in Eq.~\eqref{eq:safe-model-objective} equals $R_i(\theta)^2$, which recovers Eq.~\eqref{eq:model-objective}.
We minimize this objective with Adam.
Appendix~\ref{app:gradient} gives the explicit gradient of the ideal profiled objective.

\subsection{Estimation and inference}

Conditional on the design learned from historical data, the evaluation scores are independent and have mean $\Delta$.
We estimate the variance of their sample mean with
\begin{equation}
 \widehat{\mathrm{SE}}^2(\widehat\Delta_m)
 =\frac{1}{n(n-1)}\sum_{i=1}^{n}
 \{Z_{m,i}-\widehat\Delta_m\}^2.
 \label{eq:se}
\end{equation}
Because inference is conditional on the frozen design, this standard error does not include a separate term for the uncertainty in the historical reward model.
We test $H_0:\Delta=0$ with a two-sided Student's $t$-test.

The complete procedure is as follows:
(i)~Fit an action-wise reward model on the historical logs;
(ii)~Optimize the scalar or model-based control variate by minimizing its backoff-aware profiled objective~(Eq. \ref{eq:safe-model-objective});
(iii)~compute and freeze the optimized logger after backoff (Eq.~\ref{eq:backoff}); and
(iv)~collect the evaluation data with this logger and average the scores in Eq.~\eqref{eq:score}.

%% file: sections/experiments.tex
\section{Experimental Setup}
\label{sec:experiments}

\begin{figure*}[!p]
  \centering
  \includegraphics[trim=0bp 0bp 0bp 35bp,clip,width=0.90\textwidth]
    {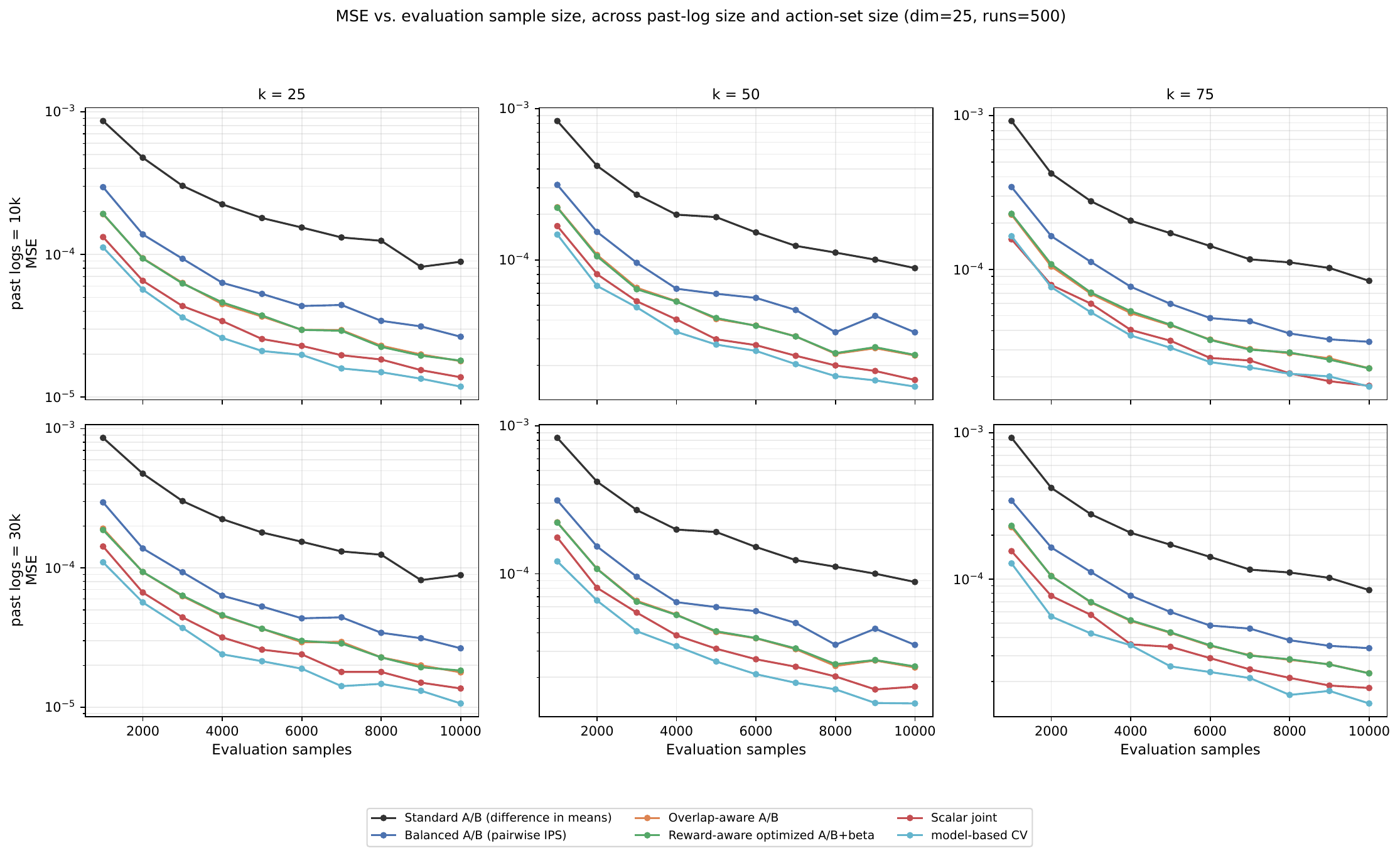}
  \caption{MSE of the ATE estimate versus the evaluation sample size $n_{\mathrm{eval}}$ (logarithmic vertical axis; lower is better).
  Columns correspond to $K=25$, 50, and 75 actions, and rows to historical logs of $n_{\mathrm{hist}}=10{,}000$ (top) and $30{,}000$ (bottom) interactions.
  Every point averages 500 independent replications with context dimension $D=25$.}
  \Description{Six-panel line chart of mean squared error versus evaluation sample size for three action counts and two historical-log sizes.
  The MSE decreases with the evaluation sample size; Scalar joint and the model-based control variate are generally below Balanced A/B and the restricted policy mixtures.}
  \label{fig:msegrid}
\end{figure*}

\begin{figure*}[!p]
  \centering
  \includegraphics[trim=0bp 0bp 0bp 35bp,clip,width=0.90\textwidth]
    {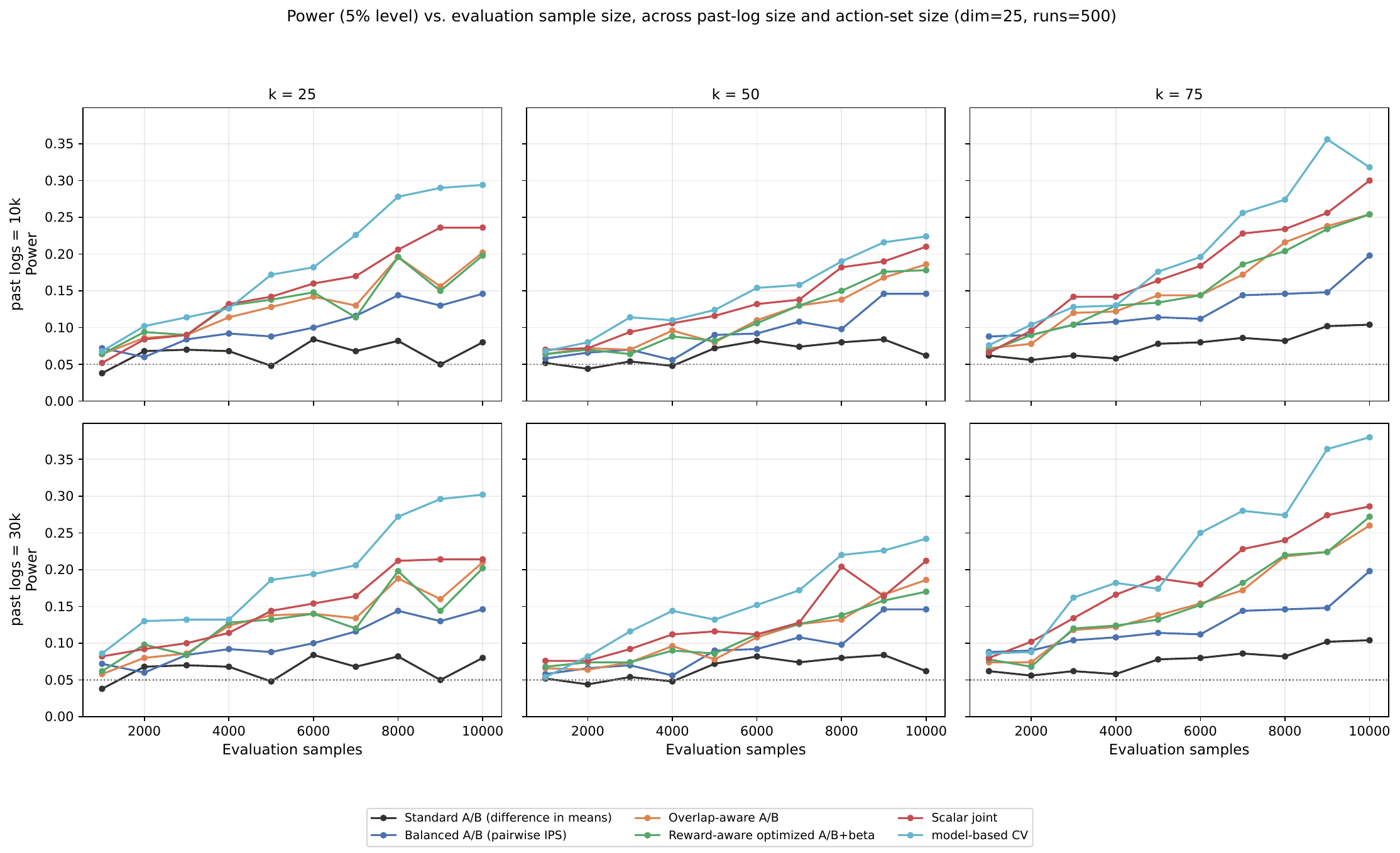}
  \caption{Empirical power of the two-sided test at the 5\% level (higher is better); the dotted line marks the 5\% level.
  The panels are arranged as in Figure~\ref{fig:msegrid}, and every point is the rejection frequency across 500 independent replications.}
  \Description{Six-panel line chart of statistical power against the evaluation sample size for three action counts and two historical-log sizes.
  Power generally increases with the evaluation sample size, and the action-level designs are usually above Balanced A/B and the restricted policy mixtures.}
  \label{fig:powergrid}
\end{figure*}

\begin{figure*}[!t]
  \centering
  \includegraphics[trim=0bp 0bp 0bp 35bp,clip,width=0.86\textwidth]
    {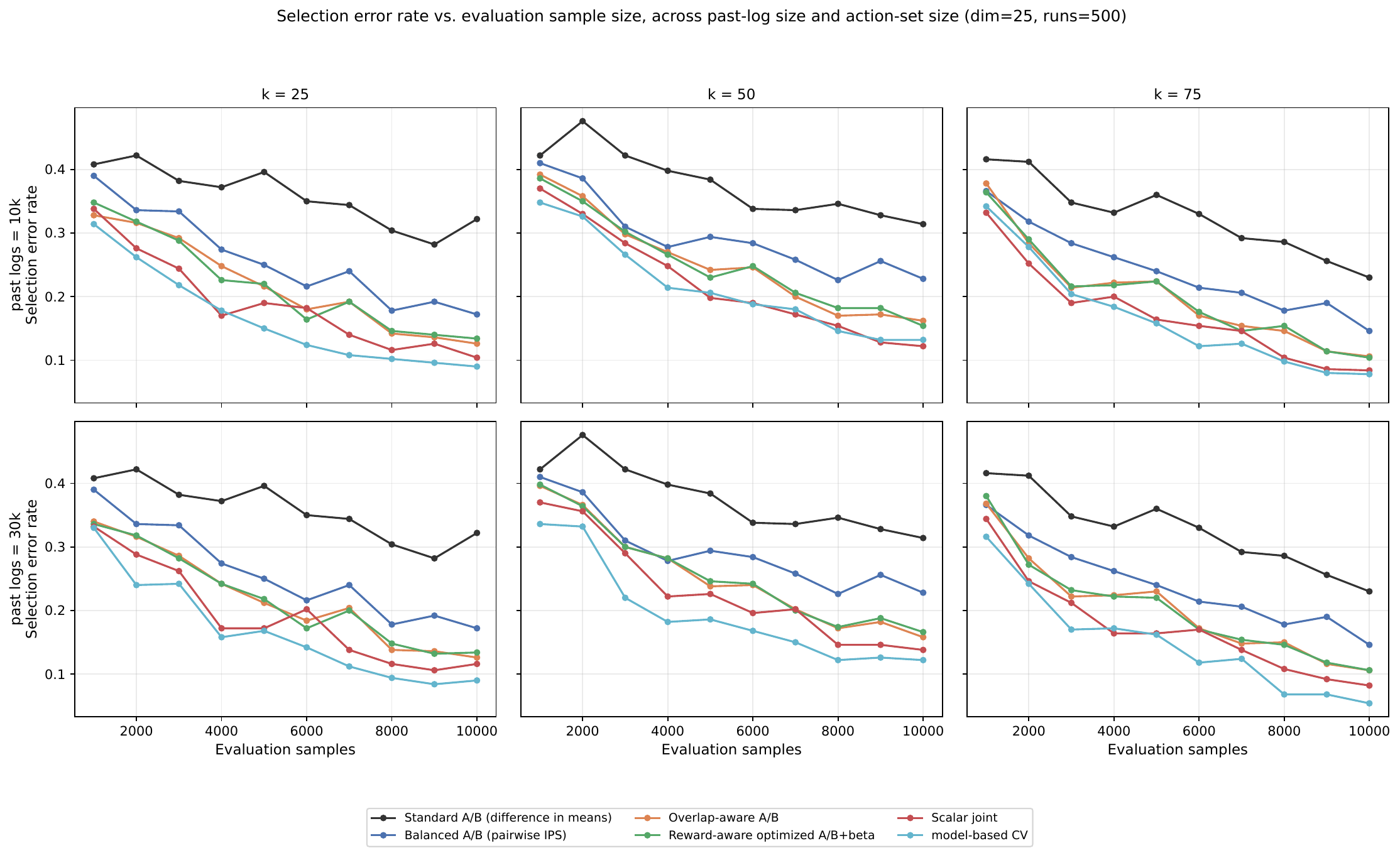}
  \caption{Wrong-sign selection rate, i.e., the frequency with which the sign of the estimate differs from that of $\Delta$, so that the worse policy would be selected (lower is better).
  The panels are arranged as in Figure~\ref{fig:msegrid}, and every point averages 500 independent replications.}
  \Description{Six-panel line chart of the wrong-sign selection rate against the evaluation sample size for three action counts and two historical-log sizes.
  The selection error generally decreases with the evaluation sample size, and the action-level methods usually have the lowest error at the larger evaluation sizes.}
  \label{fig:selectiongrid}
\end{figure*}

\subsection{Research questions}

\begin{enumerate}[label=\textbf{RQ\arabic*},leftmargin=*]
\item Does variance-optimal action-level logging improve estimation accuracy, statistical power, and policy selection over standard and balanced A/B testing and overlap-aware policy mixtures?
\item Does the model-based control variate improve over the simpler scalar joint method?
\item Are the conclusions robust to changes in action-space size, the historical log, and the evaluation sample?
\end{enumerate}

\subsection{Synthetic environment}

Our synthetic contextual-bandit environment uses Gaussian contexts ($D=25$), overlapping softmax policies, and Bernoulli rewards with context- and action-dependent means. We numerically calibrate the ATE to approximately $0.005$ for each action-space size, holding the environment fixed across sample sizes.
The historical log is collected uniformly over actions. We fit an action-wise ridge reward model and use all historical contexts to design the logger and control variate before collecting fresh evaluation data. All learned contextual designs use a $5\%$ safety backoff. 
Appendix~\ref{app:simulation} gives the exact policy and reward construction, calibration procedure, and optimization settings.

\subsection{Baselines}
\label{sec:baselines}

We compare against four practical policy-level baselines.
\emph{Standard A/B} assigns each interaction independently to the treatment or production policy with probability $1/2$ and estimates the ATE by the unadjusted difference in the two arms' mean rewards. It uses a two-sided Welch $t$-test and requires neither a historical reward model nor importance weighting.
\emph{Balanced A/B} uses $q=(\pi_t+\pi_p)/2$ and the pairwise IPS estimator in Eq.~\eqref{eq:pairwise-ips}.
It collects balanced A/B data but accounts for policy overlap, unlike the conventional difference in group means.
\emph{Overlap-aware A/B}, based on Jeunen~\cite{jeunen2026accelerating}, optimizes the global mixture $q_\rho=\rho\pi_t+(1-\rho)\pi_p$ with an overlap-aware criterion, and applies the optimal scalar control variate for this fixed logger, $\beta^*(q_\rho)$ in Eq.~\eqref{eq:beta}. \emph{Reward-aware A/B} uses the same mixture $q_\rho$, but chooses its mixing weight using the estimated reward moments as well as the policy probabilities. For each candidate $\rho$, we compute $\widehat\beta_\rho$ using Eq.~\eqref{eq:beta}, replacing $\mu$ with $\widehat\mu$ and averaging over the historical contexts. We obtain the mixing weight by numerical minimization:
\begin{equation}
 \widehat\rho=\arg\min_{\rho\in[0.01,0.99]}
 \frac{1}{n_{\mathrm{hist}}}\sum_{i=1}^{n_{\mathrm{hist}}}\sum_a
 \frac{d(x_i,a)^2\widehat s_{\widehat\beta_\rho}(x_i,a)}
 {q_\rho(a\mid x_i)},
 \label{eq:reward-aware-ab}
\end{equation}
where $\widehat s_\beta(x,a)=\widehat\mu(x,a)(1-\beta)^2+\{1-\widehat\mu(x,a)\}\beta^2$ is the estimated residual second moment for a constant control variate. This traffic split minimizes estimated variance after scalar correction but cannot set action probabilities independently of the policy mixture.
Comparing the optimized policy mixtures with Scalar joint measures the benefit of a contextual logger with a scalar control variate; comparing Scalar joint with \method{} measures the additional benefit of a context- and action-dependent control variate.

\subsection{Experiment design}

We fit the reward model and optimize the design using $n_{\mathrm{hist}}$ historical interactions and all their contexts, then estimate the ATE from $n_{\mathrm{eval}}$ fresh interactions ($n$ in Section~\ref{sec:method}).
The complete grid combines $K\in\{25,50,75\}$ actions,
$n_{\mathrm{hist}}\in\{10{,}000,30{,}000\}$ historical interactions, and
$n_{\mathrm{eval}}\in\{1{,}000,2{,}000,\ldots,10{,}000\}$ evaluation interactions,
giving 60 settings.
Each of the 500 independent replications per setting has one historical log and one evaluation batch. We freeze the designs fitted on that log before generating evaluation contexts. Methods share these contexts but sample actions from their own loggers.
To reduce Monte Carlo noise in comparisons between methods, we use common random numbers: all methods share the same evaluation contexts and the same uniform draws for action and reward sampling, while each samples actions from its own logger.
Historical and evaluation samples are generated from independent, fixed random seeds.
Standard A/B is coupled with Balanced A/B: given the sampled action $a$, we assign it to the treatment arm with probability $\pi_t(a \mid x)/\{\pi_t(a \mid x)+\pi_{p}(a \mid x)\}$. This yields exactly the distribution of a standard randomized A/B test, while sharing contexts, actions, and rewards with Balanced A/B. Standard A/B uses no historical data, so the same evaluation replications are shown in both historical-log rows.

\subsection{Metrics}

We estimate three repeated-sampling metrics across each setting's 500 independent replications.
First, the MSE is $\E[(\Deltahat-\Delta)^2]$.
Second, the empirical power is the fraction of replications in which the two-sided test at level 0.05 rejects $H_0:\Delta=0$.
Third, the selection error is
\begin{equation}
 \Pr\{\operatorname{sign}(\Deltahat)\neq\operatorname{sign}(\Delta)\},
\end{equation}
i.e., the probability of choosing the lower-reward policy based on the estimate's sign.
Lower MSE and selection error, and higher power, are better.
MSE measures estimation error, selection error records policy-order reversals, and power records whether the estimate, relative to its standard error, is far enough from zero to reject the null.

\section{Results}
\label{sec:results}

\subsection{Comparison with A/B testing and policy mixtures}
\label{sec:rq1}

\textbf{Estimation accuracy.}
Figure~\ref{fig:msegrid} reports the MSE as the evaluation sample size increases from 1{,}000 to 10{,}000.
Standard A/B has the highest MSE across the grid, followed by Balanced A/B, and Overlap-aware A/B and Reward-aware A/B have similar curves.
In contrast, Scalar joint and \method{} have a lower MSE than all four baselines in every panel and at every evaluation size.
For unbiased estimators with fixed designs, MSE is inversely proportional to evaluation sample size, so MSE ratios equal the traffic ratios needed for equal accuracy.
Averaged geometrically over the ten evaluation sizes, this ratio is 0.49 to 0.53 for Scalar joint and 0.39 to 0.51 for \method{} relative to Balanced A/B, and 0.73 to 0.77 and 0.59 to 0.74, respectively, relative to the better of the two policy mixtures.
Hence, \method{} reaches the MSE of Balanced A/B with only 39\% to 51\% of its evaluation traffic.
Relative to Standard A/B, the geometrically averaged MSE ratio of \method{} is 0.12 to 0.19, an 81\% to 88\% reduction. The improvement of Balanced A/B over Standard A/B also shows the benefit of accounting for policy overlap before optimizing the logger.

Reward-aware A/B and Scalar joint both fit a scalar control variate on historical reward predictions. Reward-aware A/B selects one mixing weight, whereas Scalar joint chooses an action distribution for every context.
The lower MSE of Scalar joint suggests that this additional freedom matters. However, the comparison does not isolate the effect of the logger because each method fits its scalar for its own allocation.

\textbf{Statistical power.}
Figure~\ref{fig:powergrid} shows generally increasing power with evaluation sample size: at 10{,}000 observations, Balanced A/B reaches roughly 0.15 to 0.20 across the six panels, whereas Scalar joint and \method{} reach 0.21 to 0.38; \method{} has the highest power in every panel, between 1.5 and 2.1 times that of Balanced A/B.
At the same evaluation size, Standard A/B has power between 0.062 and 0.104, so \method{} has 3.1 to 3.9 times the power.
The absolute power remains below 0.40 in the studied range.

\textbf{Selecting the better policy.}
Figure~\ref{fig:selectiongrid} shows that the selection error decreases as more evaluation observations are collected: with 10{,}000 evaluation observations, Balanced A/B has a wrong-sign rate of approximately 0.15 to 0.23, whereas Scalar joint and \method{} have rates of 0.08 to 0.14 and 0.05 to 0.13, respectively; that is, \method{} reduces the wrong-sign rate of Balanced A/B by 42\% to 63\%.
Standard A/B has a wrong-sign rate of 0.23 to 0.32 at this evaluation size; relative to it, \method{} reduces the wrong-sign rate by 58\% to 77\%.
Since this metric only considers the sign of the estimated effect, a correct sign does not imply statistical significance, and a low selection error can coexist with a modest power.

Thus, variance-optimal action-level logging generally improves all three metrics over Standard A/B, Balanced A/B (pairwise IPS), and the evaluated policy mixtures in our simulations (RQ1).

\subsection{Scalar versus model-based control variates}
\label{sec:rq2}

\method{} improves on Scalar joint in most cases: across the 60 combinations of panel and evaluation size, it has a lower MSE in 58, a higher power in 53, and a lower selection error in 51, and all exceptions in power occur at evaluation sizes of at most 5{,}000.
Since both methods use a contextual logger, this shows that a control variate that depends on the context and the action can help, whilst the scalar method remains competitive.

Thus, model-based correction outperforms scalar correction in most settings, particularly at larger evaluation sizes, but not uniformly; scalar correction remains competitive (RQ2).

\subsection{Robustness across settings}
\label{sec:rq3}

\textbf{Historical sample size.}
Increasing the historical sample from 10,000 to 30,000 interactions reduces the
MSE of Model-CV in 28 of the 30 combinations of action-space and evaluation size,
but has little effect on Scalar joint (13 of 30), whose design depends on the
historical data only through a single parameter and, weakly, through the reward
model in its logger.

\textbf{Number of actions.}
Performance does not worsen monotonically as $K$ increases from 25 to 75; for example, some settings with $K=75$ have lower errors than their counterparts with $K=50$.
Since each action count uses a separately generated and calibrated environment, these panels do not isolate the effect of adding actions, but show whether comparisons between methods persist across three different environments.

Within the studied range, rankings are broadly similar across action-space, historical-log, and evaluation sizes; additional evaluation data yield more consistent benefits than additional historical data (RQ3).

\textbf{Evaluation sample size.} Scalar joint and Model-CV have a lower MSE than
all four baselines at every evaluation size from 1,000 to 10,000. Differences in
power and selection error are smallest at the lowest evaluation sizes, where all
methods have power close to the 5\% level.

Thus, the advantage of action-level logging in MSE persists across action-space
sizes, historical-log sizes, and evaluation sizes, whereas its advantages in
power and selection error are clearest at larger evaluation sizes (RQ3).

%% file: sections/discussion.tex
\section{Discussion}
\label{sec:discussion}

\textbf{Interpreting the comparison.}
Overlap-aware A/B and Reward-aware A/B both choose a single mixing weight between the two policies and perform similarly in our simulations.
Scalar joint and \method{} instead choose an action distribution for every context and obtain a lower MSE.
This pattern supports using a broader class of logging policies when the platform permits it. It does not establish that every learned contextual logger outperforms a policy mixture.
The result depends on the fitted reward model, the control variate, and the optimization. Corollary~\ref{cor:dominance} holds only for a fixed control variate and the true reward moments, before the uniform backoff.

\textbf{Scalar versus model-based control variates.}
Scalar joint offers a useful starting point: its control variate has only one parameter, it performs well in our simulations, and it is inexpensive to optimize once the historical reward model has been fitted.
\method{} allows the correction to vary across contexts and actions, at the cost of additional parameters; its lower MSE and higher power in most settings justify this flexibility, but do not establish uniform superiority.
Therefore, the choice between the two methods should depend on the improvement obtained in the application and on the cost of fitting and maintaining the model.

%% file: sections/conclusion.tex
\section{Conclusion}
\label{sec:conclusion}

For estimating the average treatment effect between two stochastic policies, we have shown that, for a fixed control variate, the variance-minimizing logging policy samples every action in proportion to the absolute difference between the two policies' probabilities of that action, weighted by its residual reward uncertainty, which reduces the joint design problem to an optimization over the control variate.
Our scalar and model-based designs are learned from historical logs, remain unbiased even when the historical reward model is inaccurate, and, in our simulations, improve MSE, power, and policy-selection error over balanced A/B testing and optimized policy mixtures.

\textbf{Future directions.}
Useful extensions include designs for repeated users and for settings that restrict which actions may be sampled. Updating the logger as new data arrive would require inference that accounts for adaptive data collection. For ranked lists, we would need to specify whether an action is a whole list or an individual item, and how interactions between items affect the reward model.

%% file: sections/appendix.tex
\section{Additional Results and Reproducibility}

\subsection{Simulation and optimization details}
\label{app:simulation}

The simulation setup is as follows: contexts follow $X\sim\mathcal N(0, I_D)$ with $D=25$.
For each action $a$, the production policy has a coefficient vector $u_a\in\mathbb R^D$ and an intercept $c_a\in\mathbb R$, giving the logit $x^\top u_a+c_a$. We draw each entry of $u_a$ from $\mathcal N(0,0.35^2)$ and $c_a$ from $\mathcal N(0,0.25^2)$.
The treatment policy adds $1.1(x^\top v_a+e_a)$ to the production logit. Here, $v_a\in\mathbb R^D$ and $e_a\in\mathbb R$ describe the change in the coefficients and intercept for action $a$, with entries drawn from $\mathcal N(0,0.22^2)$ and $\mathcal N(0,0.12^2)$, respectively. Both policies apply a softmax to their logits to obtain action probabilities.
The rewards are Bernoulli with mean:
\begin{gather}
\mu(x,a)=\operatorname{sigmoid}\{x^\top r_a+h_a+\kappa Kd(x,a)\},
 \label{eq:dgp}
\end{gather}
where $r_a\in\mathbb R^D$ and $h_a\in\mathbb R$ are the reward coefficients and intercept for action $a$. We draw the entries of $r_a$ from $\mathcal N(0,0.28^2)$ and $h_a$ from $\mathcal N(-1.1,0.35^2)$.
The additional term $\kappa Kd(x,a)$ lets us set the effect size in the simulation. Here, $d(x,a)=\pi_t(a\mid x)-\pi_p(a\mid x)$ is the policy contrast from Eq.~\eqref{eq:pairwise-ate}, $K$ is the number of actions, and $\kappa$ is a scalar calibration parameter. The factor $K$ rescales the contrast, while $\kappa$ controls how strongly it changes the reward mean. This adjustment changes the rewards without changing either policy.
For each action-space size, we choose $\kappa$ so that the ATE is approximately $0.005$. We choose $\kappa$ by one-dimensional root-finding so that the ATE, approximated by quasi-Monte Carlo averaging over $2^{14}$ scrambled-Sobol Gaussian contexts, equals 0.005.
The calibration contexts are separate from the historical and evaluation data and are used only to define the environment; the resulting effect is a numerical approximation to the population ATE.
For a given number of actions, the environment parameters are fixed across all historical-log and evaluation sizes, and only the sampled historical and evaluation data vary.

In the simulation, the historical log is collected by a uniform logging policy, for simplicity.
We fit one ridge linear-probability reward model per action, with an unpenalized intercept and a ridge coefficient of 5, and clip its predictions to $[0.01,0.99]$.
We use all historical contexts to design the logger.
We initialize the logistic control variate from the action-wise mean predictions, and optimize it with 44 minibatch updates of Adam (batch size 512, learning rate 0.02, and weight decay $10^{-4}$).
Across all experiments, the context dimension $(D = 25)$, the target ATE (0.005), and the safety backoff ($\varepsilon$ = 0.05) are held fixed.

\subsection{Gradient of the profiled objective}
\label{app:gradient}

We can also fit the model-based control variate using the explicit gradient of the ideal profiled objective in Eq.~\eqref{eq:model-objective}.
Let $u_{ia}(\theta)=\nabla_\theta m_{ia}(\theta)$.
Because $\widehat\mu_{ia}$ is fixed during the optimization of the design, differentiating the quantities in Eq.~\eqref{eq:model-quantities} gives
\begin{align*}
 \nabla_\theta\widehat s_{ia}(\theta)
 &=2\{m_{ia}(\theta)-\widehat\mu_{ia}\}u_{ia}(\theta),\\
 \nabla_\theta g_i(\theta)
 &=\sum_a d_{ia}u_{ia}(\theta),\\
 \nabla_\theta R_i(\theta)
 &=\sum_a |d_{ia}|
 \frac{m_{ia}(\theta)-\widehat\mu_{ia}}{\sqrt{\widehat s_{ia}(\theta)}}u_{ia}(\theta).
\end{align*}
Differentiating every summand of Eq.~\eqref{eq:model-objective} then yields
\begin{align*}
 \nabla_\theta\widehat{\mathcal F}(\theta)
 =\frac{2}{n_{\mathrm{hist}}}\sum_{i=1}^{n_{\mathrm{hist}}}\Big[&
 R_i(\theta)\nabla_\theta R_i(\theta)\\
 &+\{\widehat\tau_i-g_i(\theta)\}\nabla_\theta g_i(\theta)\Big].
\end{align*}
For the logistic model in Eq.~\eqref{eq:sigmoid}, the gradient of the model output is
\begin{equation*}
 u_{ia}(\theta)=m_{ia}(\theta)\{1-m_{ia}(\theta)\}\nabla_\theta(w_a^\top x_i+b_a),
\end{equation*}
which is nonzero only for the parameters $(w_a,b_a)$ of action $a$.
Since $\widehat s_{ia}(\theta)=\widehat\mu_{ia}(1-\widehat\mu_{ia})+\{\widehat\mu_{ia}-m_{ia}(\theta)\}^2$ and the predictions $\widehat\mu_{ia}$ are clipped to $[0.01,0.99]$, the residual second moment is bounded away from zero. The gradient is therefore well defined.

%% file: main.bbl

\begin{thebibliography}{24}


\ifx \showCODEN    \undefined \def \showCODEN     #1{\unskip}     \fi
\ifx \showISBNx    \undefined \def \showISBNx     #1{\unskip}     \fi
\ifx \showISBNxiii \undefined \def \showISBNxiii  #1{\unskip}     \fi
\ifx \showISSN     \undefined \def \showISSN      #1{\unskip}     \fi
\ifx \showLCCN     \undefined \def \showLCCN      #1{\unskip}     \fi
\ifx \shownote     \undefined \def \shownote      #1{#1}          \fi
\ifx \showarticletitle \undefined \def \showarticletitle #1{#1}   \fi
\ifx \showURL      \undefined \def \showURL       {\relax}        \fi
\providecommand\bibfield[2]{#2}
\providecommand\bibinfo[2]{#2}
\providecommand\natexlab[1]{#1}
\providecommand\showeprint[2][]{arXiv:#2}

\bibitem[Deng et~al\mbox{.}(2013)]%
        {deng2013cuped}
\bibfield{author}{\bibinfo{person}{Alex Deng}, \bibinfo{person}{Ya Xu}, \bibinfo{person}{Ron Kohavi}, {and} \bibinfo{person}{Toby Walker}.} \bibinfo{year}{2013}\natexlab{}.
\newblock \showarticletitle{Improving the sensitivity of online controlled experiments by utilizing pre-experiment data}. In \bibinfo{booktitle}{\emph{Proceedings of the sixth ACM international conference on Web search and data mining}}. \bibinfo{pages}{123--132}.
\newblock


\bibitem[Dud{\'\i}k et~al\mbox{.}(2011)]%
        {dudik2011doubly}
\bibfield{author}{\bibinfo{person}{Miroslav Dud{\'\i}k}, \bibinfo{person}{John Langford}, {and} \bibinfo{person}{Lihong Li}.} \bibinfo{year}{2011}\natexlab{}.
\newblock \showarticletitle{Doubly robust policy evaluation and learning}.
\newblock \bibinfo{journal}{\emph{arXiv preprint arXiv:1103.4601}} (\bibinfo{year}{2011}).
\newblock


\bibitem[Farajtabar et~al\mbox{.}(2018)]%
        {farajtabar2018more}
\bibfield{author}{\bibinfo{person}{Mehrdad Farajtabar}, \bibinfo{person}{Yinlam Chow}, {and} \bibinfo{person}{Mohammad Ghavamzadeh}.} \bibinfo{year}{2018}\natexlab{}.
\newblock \showarticletitle{More robust doubly robust off-policy evaluation}. In \bibinfo{booktitle}{\emph{International Conference on Machine Learning}}. PMLR, \bibinfo{pages}{1447--1456}.
\newblock


\bibitem[Guo et~al\mbox{.}(2021)]%
        {guo2021mlrate}
\bibfield{author}{\bibinfo{person}{Yongyi Guo}, \bibinfo{person}{Dominic Coey}, \bibinfo{person}{Mikael Konutgan}, \bibinfo{person}{Wenting Li}, \bibinfo{person}{Chris Schoener}, {and} \bibinfo{person}{Matt Goldman}.} \bibinfo{year}{2021}\natexlab{}.
\newblock \showarticletitle{Machine learning for variance reduction in online experiments}.
\newblock \bibinfo{journal}{\emph{Advances in Neural Information Processing Systems}}  \bibinfo{volume}{34} (\bibinfo{year}{2021}), \bibinfo{pages}{8637--8648}.
\newblock


\bibitem[Gupta(2025)]%
        {gupta2025safe}
\bibfield{author}{\bibinfo{person}{Shashank Gupta}.} \bibinfo{year}{2025}\natexlab{}.
\newblock \showarticletitle{Safe, Efficient, and Robust Reinforcement Learning for Ranking and Diffusion Models}.
\newblock \bibinfo{journal}{\emph{arXiv preprint arXiv:2510.15429}} (\bibinfo{year}{2025}).
\newblock


\bibitem[Gupta et~al\mbox{.}(2023a)]%
        {gupta2023recent}
\bibfield{author}{\bibinfo{person}{Shashank Gupta}, \bibinfo{person}{Philipp Hager}, \bibinfo{person}{Jin Huang}, \bibinfo{person}{Ali Vardasbi}, {and} \bibinfo{person}{Harrie Oosterhuis}.} \bibinfo{year}{2023}\natexlab{a}.
\newblock \showarticletitle{Recent advances in the foundations and applications of unbiased learning to rank}. In \bibinfo{booktitle}{\emph{Proceedings of the 46th International ACM SIGIR Conference on Research and Development in Information Retrieval}}. \bibinfo{pages}{3440--3443}.
\newblock


\bibitem[Gupta et~al\mbox{.}(2024a)]%
        {gupta2024unbiased}
\bibfield{author}{\bibinfo{person}{Shashank Gupta}, \bibinfo{person}{Philipp Hager}, \bibinfo{person}{Jin Huang}, \bibinfo{person}{Ali Vardasbi}, {and} \bibinfo{person}{Harrie Oosterhuis}.} \bibinfo{year}{2024}\natexlab{a}.
\newblock \showarticletitle{Unbiased learning to rank: On recent advances and practical applications}. In \bibinfo{booktitle}{\emph{Proceedings of the 17th ACM International Conference on Web Search and Data Mining}}. \bibinfo{pages}{1118--1121}.
\newblock


\bibitem[Gupta et~al\mbox{.}(2024b)]%
        {gupta2024optimal}
\bibfield{author}{\bibinfo{person}{Shashank Gupta}, \bibinfo{person}{Olivier Jeunen}, \bibinfo{person}{Harrie Oosterhuis}, {and} \bibinfo{person}{Maarten de Rijke}.} \bibinfo{year}{2024}\natexlab{b}.
\newblock \showarticletitle{Optimal baseline corrections for off-policy contextual bandits}. In \bibinfo{booktitle}{\emph{Proceedings of the 18th ACM Conference on Recommender Systems}}. \bibinfo{pages}{722--732}.
\newblock


\bibitem[Gupta et~al\mbox{.}(2025)]%
        {gupta2025towards}
\bibfield{author}{\bibinfo{person}{Shashank Gupta}, \bibinfo{person}{Yiming Liao}, {and} \bibinfo{person}{Maarten de Rijke}.} \bibinfo{year}{2025}\natexlab{}.
\newblock \showarticletitle{Towards Two-Stage Counterfactual Learning to Rank}. In \bibinfo{booktitle}{\emph{Proceedings of the 2025 International ACM SIGIR Conference on Innovative Concepts and Theories in Information Retrieval (ICTIR)}}. \bibinfo{pages}{177--182}.
\newblock


\bibitem[Gupta et~al\mbox{.}(2023b)]%
        {gupta2023safe}
\bibfield{author}{\bibinfo{person}{Shashank Gupta}, \bibinfo{person}{Harrie Oosterhuis}, {and} \bibinfo{person}{Maarten de Rijke}.} \bibinfo{year}{2023}\natexlab{b}.
\newblock \showarticletitle{Safe deployment for counterfactual learning to rank with exposure-based risk minimization}. In \bibinfo{booktitle}{\emph{Proceedings of the 46th International ACM SIGIR Conference on Research and Development in Information Retrieval}}. \bibinfo{pages}{249--258}.
\newblock


\bibitem[Hanna et~al\mbox{.}(2017)]%
        {hanna2017data}
\bibfield{author}{\bibinfo{person}{Josiah~P Hanna}, \bibinfo{person}{Philip~S Thomas}, \bibinfo{person}{Peter Stone}, {and} \bibinfo{person}{Scott Niekum}.} \bibinfo{year}{2017}\natexlab{}.
\newblock \showarticletitle{Data-efficient policy evaluation through behavior policy search}. In \bibinfo{booktitle}{\emph{International Conference on Machine Learning}}. PMLR, \bibinfo{pages}{1394--1403}.
\newblock


\bibitem[Horvitz and Thompson(1952)]%
        {horvitz1952generalization}
\bibfield{author}{\bibinfo{person}{Daniel~G Horvitz} {and} \bibinfo{person}{Donovan~J Thompson}.} \bibinfo{year}{1952}\natexlab{}.
\newblock \showarticletitle{A generalization of sampling without replacement from a finite universe}.
\newblock \bibinfo{journal}{\emph{Journal of the American statistical Association}} \bibinfo{volume}{47}, \bibinfo{number}{260} (\bibinfo{year}{1952}), \bibinfo{pages}{663--685}.
\newblock


\bibitem[Jeunen(2026)]%
        {jeunen2026accelerating}
\bibfield{author}{\bibinfo{person}{Olivier Jeunen}.} \bibinfo{year}{2026}\natexlab{}.
\newblock \showarticletitle{Accelerating A/B-Tests with Counterfactual Estimation: Reducing Variance through Policy Overlap}.
\newblock \bibinfo{journal}{\emph{arXiv preprint arXiv:2607.14604}} (\bibinfo{year}{2026}).
\newblock


\bibitem[Jeunen and Ustimenko(2024)]%
        {jeunen2024delta}
\bibfield{author}{\bibinfo{person}{Olivier Jeunen} {and} \bibinfo{person}{Aleksei Ustimenko}.} \bibinfo{year}{2024}\natexlab{}.
\newblock \showarticletitle{$\delta$-OPE: Off-policy estimation with pairs of policies}. In \bibinfo{booktitle}{\emph{Proceedings of the 18th ACM Conference on Recommender Systems}}. \bibinfo{pages}{878--883}.
\newblock


\bibitem[Kato et~al\mbox{.}(2020)]%
        {kato2020efficient}
\bibfield{author}{\bibinfo{person}{Masahiro Kato}, \bibinfo{person}{Takuya Ishihara}, \bibinfo{person}{Junya Honda}, {and} \bibinfo{person}{Yusuke Narita}.} \bibinfo{year}{2020}\natexlab{}.
\newblock \showarticletitle{Efficient adaptive experimental design for average treatment effect estimation}.
\newblock \bibinfo{journal}{\emph{arXiv preprint arXiv:2002.05308}} (\bibinfo{year}{2020}).
\newblock


\bibitem[Kohavi et~al\mbox{.}(2020)]%
        {kohavi2020trustworthy}
\bibfield{author}{\bibinfo{person}{Ron Kohavi}, \bibinfo{person}{Diane Tang}, {and} \bibinfo{person}{Ya Xu}.} \bibinfo{year}{2020}\natexlab{}.
\newblock \bibinfo{booktitle}{\emph{Trustworthy online controlled experiments: A practical guide to a/b testing}}.
\newblock \bibinfo{publisher}{Cambridge University Press}.
\newblock


\bibitem[Konishi et~al\mbox{.}(2026)]%
        {konishi2026more}
\bibfield{author}{\bibinfo{person}{Koki Konishi}, \bibinfo{person}{Masataka Ushiku}, {and} \bibinfo{person}{Yuta Saito}.} \bibinfo{year}{2026}\natexlab{}.
\newblock \showarticletitle{A More Accurate Algorithm Comparison through A/B Testing using Offline Evaluation Methods}. In \bibinfo{booktitle}{\emph{Proceedings of the 32nd ACM SIGKDD Conference on Knowledge Discovery and Data Mining V. 2}}. \bibinfo{pages}{2401--2412}.
\newblock


\bibitem[Liu et~al\mbox{.}(2025)]%
        {liu2024doubly}
\bibfield{author}{\bibinfo{person}{Shuze Liu}, \bibinfo{person}{Claire Chen}, {and} \bibinfo{person}{Shangtong Zhang}.} \bibinfo{year}{2025}\natexlab{}.
\newblock \showarticletitle{Doubly optimal policy evaluation for reinforcement learning}. In \bibinfo{booktitle}{\emph{International Conference on Learning Representations}}, Vol.~\bibinfo{volume}{2025}. \bibinfo{pages}{10793--10820}.
\newblock


\bibitem[Liu and Zhang(2023)]%
        {liu2023efficient}
\bibfield{author}{\bibinfo{person}{Shuze Liu} {and} \bibinfo{person}{Shangtong Zhang}.} \bibinfo{year}{2023}\natexlab{}.
\newblock \showarticletitle{Efficient policy evaluation with offline data informed behavior policy design}.
\newblock \bibinfo{journal}{\emph{arXiv preprint arXiv:2301.13734}} (\bibinfo{year}{2023}).
\newblock


\bibitem[Poyarkov et~al\mbox{.}(2016)]%
        {poyarkov2016boosted}
\bibfield{author}{\bibinfo{person}{Alexey Poyarkov}, \bibinfo{person}{Alexey Drutsa}, \bibinfo{person}{Andrey Khalyavin}, \bibinfo{person}{Gleb Gusev}, {and} \bibinfo{person}{Pavel Serdyukov}.} \bibinfo{year}{2016}\natexlab{}.
\newblock \showarticletitle{Boosted decision tree regression adjustment for variance reduction in online controlled experiments}. In \bibinfo{booktitle}{\emph{Proceedings of the 22nd ACM SIGKDD International Conference on Knowledge Discovery and Data Mining}}. \bibinfo{pages}{235--244}.
\newblock


\bibitem[Sakhi et~al\mbox{.}(2026)]%
        {sakhi2026exploiting}
\bibfield{author}{\bibinfo{person}{Otmane Sakhi}, \bibinfo{person}{Alexandre Gilotte}, {and} \bibinfo{person}{David Rohde}.} \bibinfo{year}{2026}\natexlab{}.
\newblock \showarticletitle{Exploiting Similarities in A/B Testing with Off-Policy Estimation}. In \bibinfo{booktitle}{\emph{Proceedings of the 32nd ACM SIGKDD Conference on Knowledge Discovery and Data Mining V. 2}}. \bibinfo{pages}{4103--4114}.
\newblock


\bibitem[Tucker and Joachims(2023)]%
        {tucker2023variance}
\bibfield{author}{\bibinfo{person}{Aaron~D Tucker} {and} \bibinfo{person}{Thorsten Joachims}.} \bibinfo{year}{2023}\natexlab{}.
\newblock \showarticletitle{Variance-minimizing augmentation logging for counterfactual evaluation in contextual bandits}. In \bibinfo{booktitle}{\emph{Proceedings of the Sixteenth ACM International Conference on Web Search and Data Mining}}. \bibinfo{pages}{967--975}.
\newblock


\bibitem[Wan et~al\mbox{.}(2022)]%
        {wan2022safe}
\bibfield{author}{\bibinfo{person}{Runzhe Wan}, \bibinfo{person}{Branislav Kveton}, {and} \bibinfo{person}{Rui Song}.} \bibinfo{year}{2022}\natexlab{}.
\newblock \showarticletitle{Safe exploration for efficient policy evaluation and comparison}. In \bibinfo{booktitle}{\emph{International Conference on Machine Learning}}. PMLR, \bibinfo{pages}{22491--22511}.
\newblock


\bibitem[Wang et~al\mbox{.}(2017)]%
        {wang2017optimal}
\bibfield{author}{\bibinfo{person}{Yu-Xiang Wang}, \bibinfo{person}{Alekh Agarwal}, {and} \bibinfo{person}{Miroslav Dud{\i}k}.} \bibinfo{year}{2017}\natexlab{}.
\newblock \showarticletitle{Optimal and adaptive off-policy evaluation in contextual bandits}. In \bibinfo{booktitle}{\emph{International Conference on Machine Learning}}. PMLR, \bibinfo{pages}{3589--3597}.
\newblock


\end{thebibliography}
